\documentclass[11pt]{article}
\usepackage{acl2015}
\usepackage{times}
\usepackage{url}
\usepackage{latexsym}

\title{Combine CRF and MMSEG to Boost Chinese Word Segmentation in Social Media}
\author{Yushi Yao\\
	    Shanghai Jiaotong University\\
	    800 Dongchuan Street\\
	    Minhang District, Shanghai, China\\
	    {\tt yys12345@sjtu.edu.cn}
	  \And
	Zheng Huang\\
  	Shanghai Jiaotong University\\
  	800 Dongchuan Street\\
  	Minhang District,Shanghai , China\\
  {\tt huangzhengsjtu@126.com}}
  
\date{}

\begin{document}
\maketitle
\begin{abstract}
  In this paper, we propose a joint algorithm for the word segmentation on Chinese social media. Previous work mainly focus on word segmentation for plain Chinese text, in order to develop a Chinese social media processing tool, we need to take the main features of social media into account, whose grammatical structure is not rigorous, and the tendency of using colloquial and Internet terms makes the existing Chinese-processing tools inefficient to obtain good performance on social media.~\cite{Ronan:11}
  In our approach, we combine CRF and MMSEG algorithm and extend features of traditional CRF algorithm to train the model for word segmentation, We use Internet lexicon in order to improve the performance of our model on Chinese social media. Our experimental result on Sina Weibo shows that our approach outperforms the state-of-the-art model.
\end{abstract}

\section{Introduction}

  Social media contains vast amounts of information, all of the information can be used to explore micro-blog event and apply sentiment analysis, etc. In order to obtain essential information, first step is text processing, generally we use natural language processing methods to analyze social media and extract information for other social media research.~\cite{Xiong:13}

  An English sentence uses spaces as the gap between words, but Chinese has no such word boundaries, so Chinese natural language process includes one more step than English: Word Segmentation -- determination of the word boundaries. Word Segmentation is the core step of the entire Chinese language processing, it is directly related to the accuracy of the entire Chinese natural language process, in addition, the difficulty of Chinese word segmentation lies in the removal of ambiguity and identifying OOV words. Ambiguity means that a sentence may result in many possible word segmentation results and every kind of segmentation has different semantic meanings, while the meaning of OOV words identification refers to the word which is not included by word dictionary, the most typical example is person's name, place name and so on.

\section{Related Work}

  Currently research on natural language process for social media is still at the starting stage, in the 2006 SIGHAN Chinese word segmentation competition, approaches based on sequence annotation has been widely used. Microsoft used conditional random field and the size of feature window equals one.~\cite{Huang:07} It turns out that features have been simplified, but the performance was still very good. DaLian Science and Technology University built two models, one is based on the use of the word CRF model, the other is based on MMSM model. During our experiments, we used CRF algorithm as well, and it plays the key role when we build the model. Our work mainly depends on the improvement of conditional random field.

  Conditional random field (CRF) is a statistical sequence modeling framework which is first introduced into language processing by Lafferty J et al(2001).~\cite{Lafferty:01} Previous research showed that CRF has a good performance on word segmentation accuracy in the pipeline method. Tseng H et al(2005) and John Lafferty et al(2001) introduced a conditional random field sequence model in conjunction with character identity features, morphological features and character reduplication features.~\cite{Tseng:05,John:01} The study that is closely related to ours is ~\cite{Zhao:06}, which also used assistant algorithm and added external lexicon, while they just add the output of assistant algorithm to the feature templates. Different from that work, we take not only the relevance between the character and its MMSEG output tag, but also the context feature of these MMSEG output tags as well.

\section{Methods}

  There are already a lot of sophisticated algorithms for word segmentation such as: statistical methods(Hidden Markov Model HMM, CRF etc.), lexicon-based algorithms (MMSEG), and rule-based algorithms. CRF performs better than lexicon-based models on OOV rates because CRF introduce additional features,~\cite{Michael:02} which may be artificially added~\cite{Xiong:09}, including character level features and context level features, in addition, CRF also maintains the Markov characteristics of the word,~\cite{Wallach:04} thus we can remove word ambiguity by combining more features as well.

  Generally there are several simple algorithms used in word segmentation. When applying lexicon-based algorithm such as MMSEG, we simply match words according to our lexicon. While for statistical-based algorithm such as CRF, the training set is turned into a Chinese character sequence and the segmentation task can be considered as an annotation task.

  More specifically, the model trained by CRF assigns each character a word boundary tag when labeling a sentence. Here we use the BMES tag set, B, M and E denote the first, middle and last character of a multi-character word respectively, and S denotes the single character word.~\cite{McCallum:10}

  As social media contains a large amount of OOV words, and with respect to out-of-vocabulary recognition, lexicon-based algorithm has a strong advantage to identify the correct rate, so when provided a suitable lexicon, lexicon-based algorithm can be combined with CRF algorithm to enhance the performance of CRF in word segmentation of social media.~\cite{Qian:10}

  Finally we choose to use MMSEG to do rough segmentation and take the segmentation result as a new feature during the training process of the CRF segmentation model.

\section{Experiments}

  The files involved in the training process include the entire feature template file, training set and lexicon files, in order to get the best results, we adopted a similar approach as coordinates rise, we will divide the experiment into three stages, each stage select one training material to optimize and fix the other two materials.~\cite{Razvan:08} At each stage, if one training material gets the best performance on test set, it will be chosen to be the final training material that we would use.~\cite{Wallach:02}

\subsection{Data and Tool}

  The training set of word segmentation is from Backoff 2005.~\cite{Tseng:05} We have to mention that the lexicon used by MMSEG is the Sougou lab internet lexicon published in 2006, which contains a number of high-frequency words under internet environment, and mmseg4j project lexicon file.

  There are few test sets about social media, so we crawled raw content from Sina weibo and then annotate manually, the contents is from all kinds of areas and users. The size of test set is about 190K, containing about 25 thousands Chinese words. What is worth mention is that our segmentation is the same as MSR's standard.

  During our experiment, we use PCRF to train the segmentation model and test the model on our test sets, PCRF is an open source implementation of (Linear) Conditional Random Fields (CRFs) for segmenting/labeling sequential data. The train file format and feature template of PCRF are compatible with popular CRF implementations, like CRF++.

\subsection{Evaluation Metric}

  We evaluate system performance on the individual tasks. For word segmentation, three metrics are used for evaluation: precision (P), recall (R), and F-score (F) defined by 2PR/(P+R), where precision is the percentage of correct words in the system output, and recall is the percentage of words in gold standard annotations that are correctly predicted.

\subsection{Feature Template}

  \begin{table*}[t]
  \begin{center}
  \begin{tabular}{c|c|c|c}
  \hline Level & Type &  Feature & Function \\ \hline
  Character & Unigram & $C_-1$, $C_0$, $C_1$ & The previous, current, and next character \\
  Character & Bigram & $C_-1C_0$, $C_0C_1$ & The previous (next) and current characters \\
  Character & Jump & $C_-1$, $C_1$ & The previous and next character \\
  Tag & Unigram & $T_-1$, $T_0$, $T_1$ & The previous, current, and next tag \\
  Tag & Bigram & $T_-1T_0$, $T_0T_1$ & The previous (next) and current tag \\
  Tag & Jump & $T_-1$, $T_1$ & The previous and next tag \\
  Character-Tag & Bigram & $C_-1T_0$, $C_0T_0$, $C_1T_0$ & The previous, current, next character and current tag\\
  \hline
  \end{tabular}
  \end{center}
  \caption{\label{font-table} Feature Templates }
  \end{table*}

  Template file plays the core role in model training. Since CRF is designed to calculate the conditional probability of the sequence annotation deduced by the observed sequence, and this probability can be used to help describe a set of random variables related to the distribution of feature vectors, which are composed by characteristic function derived from feature templates defined by user.~\cite{Pietra:97}

  Each line in the template file of PCRF denotes one template. In each template, special macro \%x[row,col] will be used to specify a token in the input data, row specifies the relative position from the current focusing token and col specifies the absolute position of the column.

  The first stage of the experiment is to determine the feature vectors of our model,~\cite{Jenny:08} the original features are too simple so we change the feature template step by step and test the accuracy on our test set. Table 1 presents the features that get the best performance during our experiments. Our template are composed by three-level feature templates: Character-level, Tag-level and Character-Tag level, the three levels of features introduce the correlation of characters and tags respectively, as well as the correlation of the character and its neighbour tags.

  \begin{itemize}
  \item Experiment 1: No new correlation, just the correlation between each character and its neighbours.
  \item Experiment 2: New correlation between the MMSEG tag of each character, and the correlation between each MMSEG tag and its neighbour.
  \item Experiment 3: New trigram correlation among current character, its neighbour and its corresponding MMSEG tag.
  \item Experiment 4: Breaking one trigram correlation in experiment 3 into three bigram correlations, new correlation between the current character's neighbour and the MMSEG tag of current character
  \item Experiment 5: New correlation between the previous(next) two or three character, and the current MMSEG tag
  \end{itemize}

  \begin{table}[h]
  \footnotesize
  \begin{center}
  \begin{tabular}{c|cc}
  \hline  No. &  Number of Templates & \bf P { }{ }{ }{ }{ }{ } R { }{ }{ }{ }{ }{ } F \\ \hline
  1 & 6 & 96.30 {} 96.47 {} 96.39 \\
  2 & 12 & 97.14 {} 97.04 {} 97.09 \\
  3 & 16 & 97.27 {} 97.16 {} 97.21 \\
  4 & 15 & 97.28 {} 97.25 {} 97.26 \\
  5 & 21 & 96.93 {} 96.86 {} 96.89 \\
  \hline
  \end{tabular}
  \end{center}
  \caption{\label{font-table} Result of word segmentation experiments with different feature template files. }
  \end{table}

  Table2 shows the results of word segmentation experiments with different feature template files. We found that maybe more feature templates means higher precision and recall. After these experiments we consider adding more correlations, including expanding the width of correlation window to 2 or 3(Experiment 5), but the test results got setback, which indicates that our word segmentation model only need to consider unigram Markov property. Considering too much correlation will have opposite effect on the performance.

  In the segmentation experiment, the model raised nearly 0.9 percent on both precision and recall. Finally we decided to use segmentation model with the highest F-score.

\subsection{Training Set}

  The next stage is to change the training set and find out the influence of the training set on our model. We choose the training sets from MSR and PKU because the text of these two training sets are mainly from newspaper, which is similar to the content of some social media because many social medias are informative, these social medias act like news, the only difference is that they are from Internet and general news is mainly from newspapers.

  \begin{table}[h]
  \footnotesize
  \begin{center}
  \begin{tabular} {c|cc}
  \hline  No. &  Training Set & P  { }{ }{ }{ }{ }{ }  R  { }{ }{ }{ }{ }{ }  F \\ \hline
  1 & MSR & 97.28 {}  97.25 {}  97.26 \\
  2 & PKU & 88.31 {}  84.65 {}  86.44 \\
  3 & MSR+PKU & 95.12 {} 93.37 {} 94.24 \\
  \hline
  \end{tabular}
  \end{center}
  \caption{\label{font-table} Result of word segmentation experiments with different training sets. }
  \end{table}

  Table3 shows the results of word segmentation experiments with different training sets. From the result of experiments we can see that the model trained by the combination of these two training sets did not outperform the model of single training set and the F score got declined conversely, it is confusing indeed, and the main reason is the different segmentation standard of the two training set result in the fact that misleading during the training process, which will be discussed later.

\subsection{Lexicon}

  The final stage of our experiment is to choose an appropriate lexicon. We have known that the more words the lexicon contains, the higher ability of distinguishing words that MMSEG has, since our goal is to get high performance on the social media, we should add the Internet lexicon to our original lexicon, and we conducted one more experiment during which we add a lexicon of a particular field just for comparison.

  \begin{table}[h]
  \footnotesize
  \begin{center}
  \begin{tabular} {c|cc}
  \hline  No. &  Lexicon & P  { }{ }{ }{ }{ }{ }  R  { }{ }{ }{ }{ }{ }  F \\ \hline
  1 & MMSEG4J & 97.28 { } 97.25	{ } 97.26 \\
  2 & MMSEG4J+Sougou Internet & 97.28 { } 97.39 { } 97.35 \\
  3 & MMSEG4J+Sougou Medicine  & 97.28 { } 97.26 { } 97.27 \\
  \hline
  \end{tabular}
  \end{center}
  \caption{\label{font-table} Result of word segmentation experiments using different lexicons. }
  \end{table}

  Table 4 shows the results of word segmentation experiments using different lexicons. The combination of the Internet lexicon has improved the overall performance of word segmentation about 0.1\%, but if we just add the the lexicon of a particular field, such as the lexicon of medical science, it will have just a little positive effect on the test set, which is mainly due to the amount of overall coverage of terminology in specific areas is not high enough in our test set.

\subsection{Result Comparison}

  For reference we took a comparison between our best experiment result and other segmentation tools. We choose the model trained by backoff2005 MSR training set and the internet lexicon from Sougou Lab for comparison.

  \begin{table}[h]
  \footnotesize
  \begin{center}
  \begin{tabular} {c|cc}
  \hline  Segmentaion & P  { }{ }{ }{ }{ }{ }  R  { }{ }{ }{ }{ }{ }  F & Closed Beta\\ \hline
  ICTCLAS & 90.04 {} 86.28 {} 88.12 & 98.45 \\
  Stf segmenter & 85.86 {} 82.21 {} 83.995 & -\\
  LTP-Cloud & 97.23 {} 97.04 {} 97.24 & 97.24 \\
  Ours & 97.28 {} 97.25 {} 97.26 & -\\
  \hline
  \end{tabular}
  \end{center}
  \caption{\label{font-table} Result of word segmentation experiments with different training sets. }
  \end{table}

  We add a column of closed test results for each tool to the table, LTP-cloud got the second rank in the the Chinese social media segmentation evaluation task in CLP 2012, indicating that the results of its closed test has great authority for social media NLP tasks. These data just represent the performance on a small test set. Our model may have a larger gap between these sophisticated tools on bigger test sets, but the results still guarantees the good adaptability of our model on Chinese social media.

\subsection{Error Analysis}

  The most surprising result of our experiments is the combination of MSR and PKU training sets. We compared the training sets of MSR and PKU and finally get the effect of different segmentation standard on the model.

  We find there are many fundamental differences between the standards. For example, ¡°6ÔÂ13ÈÕ¡± is one word in the MSR training set, but is broken into ¡°6ÔÂ¡± and ¡°13ÈÕ¡± as two words in pku's. In addition, people's name is also quite a big issue, MSR tends to treat the whole name as a word while PKU takes the first name as a word and the last name as another one. All in all, MSR generates more words than PKU.

\section{Conclusion}

  In this paper, our research is mainly focused on the study of word segmentation for Chinese social media. One of the most applicable algorithm is the conditional random field, which is widely used in word segmentation, part of speech tagging and named entity recognition and other aspects. CRF has a large advantage in the labeling issue, and MMSEG algorithm has a comparative advantage over other algorithms on in-vocabulary words, so we use MMSEG to do segmentation first, and use the result of the MMSEG segmentation as a new feature of CRF, in the experiments, we also continue to adjust the template file and exploit more correlation between Chinese characters, after several times of adjustment we got about 1.2\% improvement on precision and recall over single CRF model.


\end{document}